\documentclass[]{article}
\usepackage[letterpaper]{geometry}
\usepackage{amta2022}
\usepackage{times}
\usepackage{url}
\usepackage{latexsym}
\usepackage{natbib}
\usepackage{layout}
\usepackage{subfig}
\usepackage{float}

\usepackage[dvipsnames]{xcolor}

\usepackage{times}
\usepackage{latexsym}

\usepackage{comment}

\usepackage{multirow}
\usepackage{multicol}
\usepackage{booktabs}
\usepackage{graphicx}
\usepackage{caption}
\usepackage{xcolor}
\usepackage{hyperref}

\usepackage{microtype}

\usepackage{times}
\usepackage{latexsym}

\usepackage{booktabs}
\usepackage[T1]{fontenc}

\usepackage[utf8]{inputenc}

\usepackage{microtype}

\begin{document}

\title{\bf How Effective is Byte Pair Encoding for Out-Of-Vocabulary Words 

in Neural Machine Translation?}

\author{\name{\bf Ali Araabi} \hfill  \addr{a.araabi@uva.nl}\\
       \name{\bf Christof Monz} \hfill \addr{c.monz@uva.nl}\\
       \name{\bf Vlad Niculae} \hfill \addr{v.niculae@uva.nl}\\
        \addr{Informatics Institute, University of Amsterdam,
        Amsterdam, The Netherlands}
}

\maketitle
\pagestyle{empty}

\begin{abstract}
	\vspace{0.5cm}
	Neural Machine Translation~(NMT) is an open vocabulary problem. As a result, dealing with the words not occurring during training~(a.k.a. out-of-vocabulary~(OOV) words) have long been a fundamental challenge for NMT systems. The predominant method to tackle this problem is Byte Pair Encoding~(BPE) which splits words, including OOV words, into sub-word segments. BPE has achieved impressive results for a wide range of translation tasks in terms of automatic evaluation metrics. While it is often assumed that by using BPE, NMT systems are capable of handling OOV words, the effectiveness of BPE in translating OOV words has not been explicitly measured. In this paper, we study to what extent BPE is successful in translating OOV words at the word-level. We analyze the translation quality of OOV words based on word type, number of segments, cross-attention weights, and the frequency of segment n-grams in the training data. Our experiments show that while careful BPE settings seem to be fairly useful in translating OOV words across datasets, a considerable percentage of OOV words are translated incorrectly. Furthermore, we highlight the slightly higher effectiveness of BPE in translating OOV words for special cases, such as named-entities and when the languages involved are linguistically close to each other.
\end{abstract}

\section{Introduction}
One of the key challenges of neural machine translation~(NMT)~\citep{SutskeverVL14,BahdanauCB14} is vocabulary sparsity; irrespective of the amount of data available for training. 
As a consequence, all NMT models suffer from out-of-vocabulary~(OOV) words.
Accordingly, a significant proportion of sentences in the test set have OOV words.
Even with millions of sentence pairs in training data,\footnote{Russian-English from WMT} 15\% of test sentences contain OOV words, while with limited training data,\footnote{Kazakh-English from WMT} OOV words appear in more than 60\% of the test sentences.

Earlier approaches to tackling the OOV problem include using a very large vocabulary~\citep{JeanCMB15}, backing off to a dictionary look-up~\citep{LuongSLVZ15}, and copying OOV words from source to the target sentence~\citep{GulcehreANZB16}. 
However, most recent approaches are based on splitting the words into smaller units and can be divided into: language-specific approaches~\citep{SmitVGK14,HuckRF17}, language-agnostic approaches~\citep{SennrichHB16a,Kudo18,KudoR18,Costa-JussaF16,CherryFBFM18}, and hybrid approaches~\citep{HuckRF17,banerjee2018meaningless,agglutinative} which inject linguistic information into language-agnostic methods.


Nowadays, the mainstream approach to address the open-vocabulary challenge in the context of NMT is Byte Pair Encoding~\citep[BPE][]{SennrichHB16a}, due to its simplicity, applicability to a wide range of languages, and high performance in terms of automatic evaluation metrics.
BPE incrementally merges the frequent bigrams such that it keeps the most frequent words intact while splitting the rare ones into multiple segments and the granularity of these subword units is controlled by a hyperparameter. 
It is often assumed that by using BPE, NMT systems are capable of handling OOV words~\footnote{Throughout the paper, OOV refers to an actual non-segmented out-of-vocabulary word, unless otherwise stated.}, since it represents them as a sequence of subword units~\citep{SennrichHB16a,WuSCLNMKCGMKSJL16} and as a result there are very few unseen tokens in the test set thereby implying that the OOV problem has been almost solved~\citep{HuckRF17,banerjee2018meaningless,LiuXWF19,LuoYYCA19,HuLMQH20}.

Previous approaches only analyze and compare BPE and/or other segmentation strategies based on their effect on the overall translation performance~\citep{HuckRF17,Kudo18,Galle19,ProvilkovEV20,HeHN20}.
To the best of our knowledge, there is no study to investigate whether BPE solves the OOV problem at the word-level.
In this paper, we aim to explore 1) to what extent OOV words still hurt the translation quality when using BPE, 2) how useful is BPE in translating different OOV types, and 3) three potential factors that improve the translation of BPE-segmented OOV words.

We first explore the translation quality of sentences containing OOV words, showing the negative effect of the presence of OOV words on translation quality, while all of them are segmented into subword units.
We further examine the translation quality of different types of OOV words, showing the improved ability of BPE in translating named entities for linguistically close language pairs, compared to moderate to relatively poor translation quality for other types of OOV words.
We also show that OOV words that received strong cross-attention weights, have high translation qualities.
Next, we explore how the granularity of segments impacts the translation quality of OOV words.
Finally, we show that there is no evidence to support the positive correlation between the translation quality of an OOV word and occurrences of its n-grams in the training set.

\section{Experimental setup}
\paragraph{Datasets}
We use German-English, Russian-English, and Romanian-English as language pairs for our experiments.
The main reasons to select these languages are twofold: the data sizes are large enough to eliminate the effect of the amount of data on translation quality, especially for rare words. 
Also, since two common types of OOVs are inflected and compound words in general, we choose Russian, Romanian, and German with varying degrees of morphology and compound words and as a representative of Slavic, Romance, and Germanic languages, respectively. 
For the Russian-English direction, we use the Yandex corpus, Common Crawl, News Commentary, and Wiki Titles from WMT2020.
We preprocess the data by limiting the length of the sentences to 200 tokens and removing sentence pairs with a source/target length ratio exceeding $1.5$, following previous work~\citep{NgYBOAE19}. 
We use the concatenation of newstest$2017$, newstest$2018$, and newstest$2019$ for evaluation.
As German-English training set we use Europarl, Common Crawl, and News Commentary from WMT$2017$ and for the test set we use newstest$2014$.
Also for Romanian-English, we use all available training data from WMT$2016$ and newstest$2016$ for evaluation purposes.
The data prepossessing pipeline for German-English and Russian-English is similar to Russian-English.
We end up with $2.64$M, $3.95$M, and $612$K training sentences and  $1078$ / $1363$, $1830$ / $2411$, and $926$ / $1385$ OOV word types / tokens for Russian-English, German-English, and Romanian-English, respectively. 
In order to obtain sub-word segmentations, we train a joint BPE model for German-English and Romanian-English and we train a BPE model separately for the Russian-English as suggested by~\cite{NgYBOAE19}.
The number of BPE merge operations is reported for different experiments in later sections.

\paragraph{Translation model} We use the Fairseq~\footnote{https://github.com/pytorch/fairseq} NMT system to train the transformer-base model.
Since we are dealing with large enough training data, it is not essential to tune the hyper-parameters~\citep{AraabiM20} and we stick to the default set of parameters reported in the original transformer paper~\citep{VaswaniSPUJGKP17}. 

\section{Data annotation}
\label{sec:2.1}
In order to analyze BPE usability for different OOV words, we randomly sample $400$ unique OOV words from the set of all OOV words for each language.
First, we manually label OOV types. For this annotation process, we employ a highly qualified native annotator for each language. It is worthwhile mentioning that one given OOV word may belong to more than one category. Then, we extract their corresponding translation from the reference sentence. 
Below, we explain how we extract the translations of OOV words from the hypothesis sentences and also how we assign quality labels to them in more details .

\paragraph{Translation of OOV words}
In order to obtain the word-level translation correspondences of NMT output, one naive approach is to use statistical word alignments~\citep{DyerCS13}.
However, their accuracy for OOV words is poor, due to the very low frequency of OOV words.
Inspired by \cite{GargPNP19} and \cite{ChenLCJL20}, we use the average wights over heads of the encoder-decoder cross-attention in the penultimate layer of the transformer to obtain the corresponding output word for a given OOV word based on the maximum attention and then manually double-check the results. 
Also, in order to find the ground truth translations of the OOV words, we manually inspect the reference sentences to extract the corresponding words.

\paragraph{Translation label}
In order to measure the translation quality of OOV words, we make use of adequacy and fluency~\citep{KoehnM06} as assessment criteria.
Given an OOV word, we manually assign one of the following three labels to its translation:
\begin{itemize}
    \item \emph{Correct}: when the translation is the exact same word or a synonym of the ground truth, such that when replaced in the reference, it does not hurt the fluency nor adequacy. For example, ``throat inflammation'' is acceptable for ``laryngitis''. 
    \item \emph{Partly correct}: when the translation only hurts either adequacy or fluency of the sentence, but not both. For example, when the translation needs a small morphological change to be considered correct, e.g., ``reserves'' instead of ``reserve''. Also, a single spelling error which is most likely to happen in named entities or technical words, falls under this category. While we acknowledge that some translations labeled ``partly correct'' might be factually wrong in possibly harmful ways, we choose to be lenient in the annotation, as NMT systems are susceptible to make such mistakes.
    \item \emph{Wrong}: this translation hurts the adequacy and fluency of the sentence such as addition, omission, or miss-translation of the word or any part of it, e.g., when the model generates ``donated'' instead of ``imposing''.
\end{itemize}

 
\section{How does BPE segmentation benefit OOV words?}

With the experimental setup described above, we now focus on answering our research questions.
In this section, we first explore lack of which n-gram is responsible for OOV creation.
Next, we measure to what extent the presence of OOVs impacts translation quality, while practically there is no unknown sub-word token when using BPE.
Then, we see how translation quality differs for various OOV types.
Finally, we investigate three potential factors responsible for different translation quality of OOV words.
\begin{figure}
    \centering
    \includegraphics[width=0.6\linewidth]{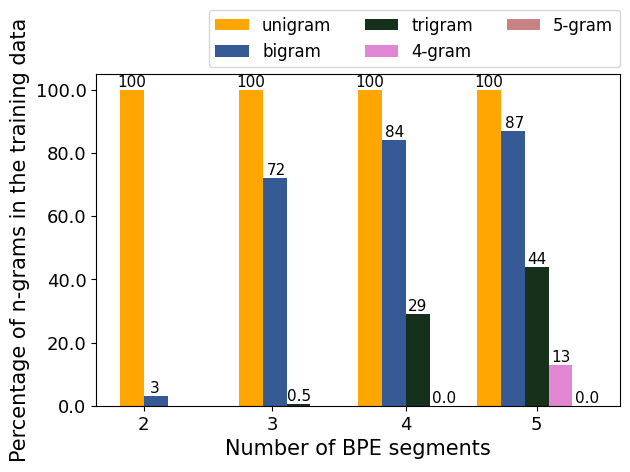}
    \caption{Presence of n-grams of BPE segmented OOV in the De-En training set for OOV words with different number of segments.}
    \label{fig:fig2}
\end{figure}

\subsection{OOV words in BPE segmented data}
Before evaluating how OOV words affect translation quality, we explore 
which n-grams of the sequence of BPE segments in the training data are responsible for creation of an OOV word at inference time.
In Figure~\ref{fig:fig2}, the horizontal axis shows German OOV words with different BPE lengths and the vertical axis shows the percentage of their n-grams present in the training set.
For example, given OOV words with three segments, obviously all of the unigrams are present in the training set, only $72$\% of their bigrams, and interestingly $0.5$\% of their trigrams are present in the training set as part of words that have more BPE segments.  
We observe that as the number of BPE segments increases, the presence of BPE n-grams in the training data increases as well. 
For various lengths of BPE segmented words, unigrams and bigrams of BPE segments are very frequent in the training data, while the longer n-grams of BPE segmentes are less frequent. 
Therefore, we conclude that lack of presence of longer sequences of OOV n-grams in the training set are responsible for OOV occurrences, while shorter n-grams are seen in the training data with a higher rate.


\subsection{The effect of OOV words on translation quality}
Translation quality can be measured either by automatic evaluation metrics such as BLEU or by human assessments.
While it has been shown that automatic MT evaluations usually fall short of human assessments~\citep{Callison-Burch09,GrahamMB14}, NMT system development has mainly focused on improving automatic evaluation metrics. 
Therefore, we use both Direct Assessment~\citep[DA][]{GrahamBMZ13} as a strong human evaluation score as well as the BLEU score to see to what extent the translation quality is affected by OOV words when BPE is applied.
In Direct Assessment, sentences are assigned a score between zero and 100 based on how adequately they express the meaning of the corresponding reference.

\begin{table}[t]
	\centering
	\begin{tabular}{lcccc}
		\toprule
	 
		\#OOV & 0 & 1 & 2 & $\geq$3 \\
		\midrule
	    Kazakh &78 & 74 &	70 & 69  \\
	    Russian &92 & 90 &	80 & 62  \\
			
		\bottomrule			
	\end{tabular}
	\caption{Median of direct assessment scores  for sentences containing various number of OOV words in Kazakh-English (low-resource) and Russian-English (high-resource). Since the scores are not normally distributed, we use the median. The higher the better.}
	\label{tab1}
\end{table}

We download the available DA scores of TALP-UPC's submission~\citep{CasasFEBC19} and Facebook FAIR's submission~\citep{NgYBOAE19} to WMT19~\footnote{www.statmt.org/wmt19/} for the Kazakh-English and Russian-English translation tasks, respectively.
The choice of these language pairs is on the grounds that we require well-performing systems trained on BPE segmented data together with their available DA scores.
Besides, we select Russian-English as a high-resource regime and Kazakh-English to represent a low-resource setting.
Table~\ref{tab1} shows the median of DA scores for sentences containing various number of OOV words in Kazakh and Russian.
In spite of using BPE which ensures almost no unknown tokens at inference time, translation quality still suffers from actual OOV words which existed before applying BPE segmentation.
In~particular, we observe that as the number of OOV words increases in a sentence, the DA score drops.
This holds for both languages, where Kazakh is considered a low-resource language and Russian as a high-resource language.
This implies that there is an inverse relationship between translation quality and the number of OOV words. Therefore, although there are no OOV words in the test sentences after applying BPE, the translation quality is lower for sentences that contain more OOV words in the absence of BPE.

\begin{figure}[]
    \centering
    \includegraphics[width=0.7\linewidth]{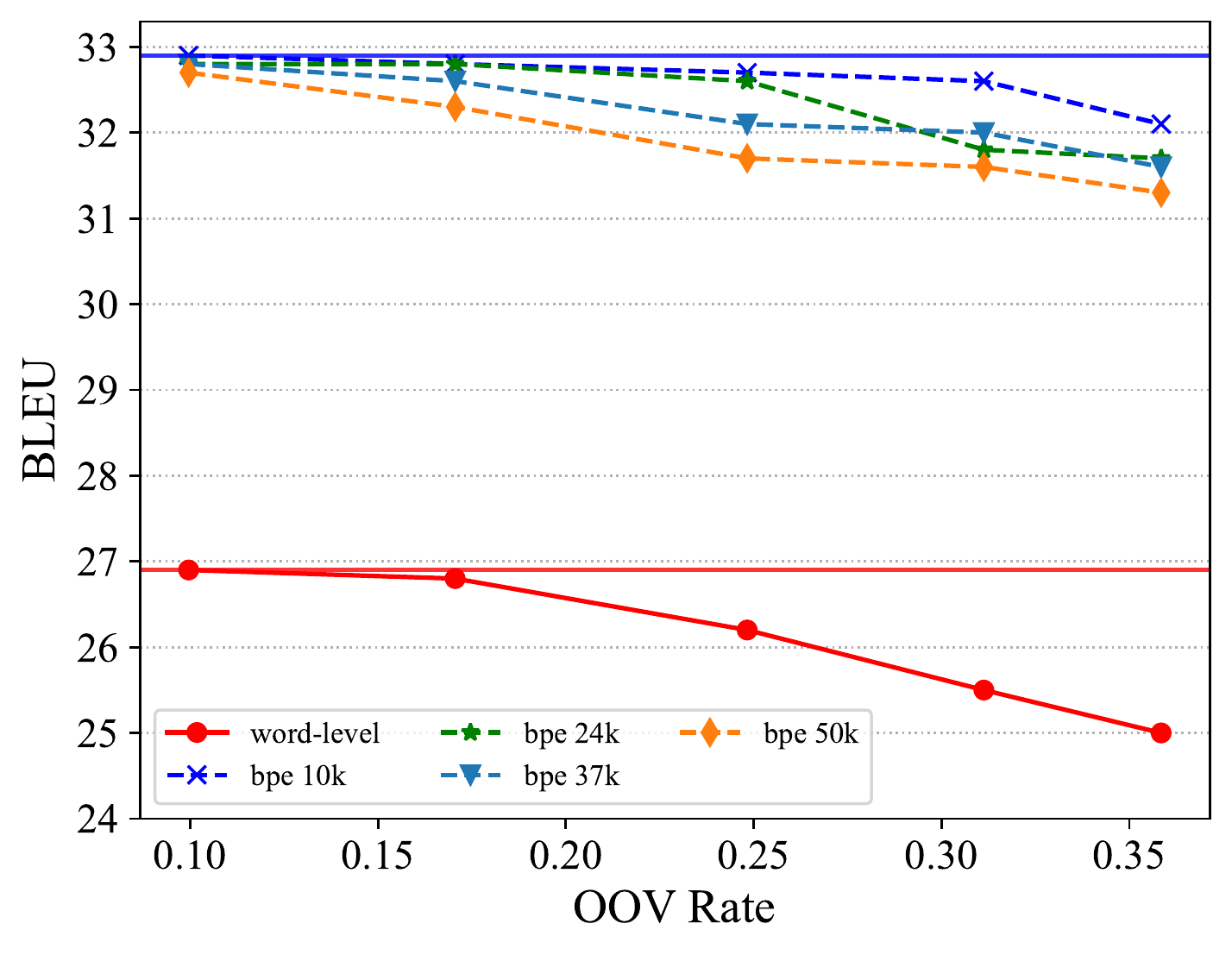}
    \caption{Comparing performance drop in word-level and BPE-level systems with different number of merge operations with increasing OOV rate in the Romanian-English test set. The horizontal lines show the performance of baselines without added OOV words.}
    \label{fig:fig5}
\end{figure}
To investigate the effect of OOV words on translation quality from the BLEU's point of view when using BPE, we take the Romanian-English dataset and add more OOV words to the test set.
In particular, we remove the least frequent words occurring in the training set---that have also occurred in the test set---from the training set by replacing them with ``<unk>'' token.
It should be noted that having high rates of OOV words (ratio of number of OOV types to the vocabulary of test set) is a realistic scenario.
Given the Romanian training set with $612$K sentences, newstest2016 has an OOV rate of $10\%$ and the Romanian test set from the Flores-101~\citep{abs-2106-03193} as a NMT benchmark has $42\%$ OOV rate.
Also, for the Kazakh-English training set with $100$K samples, the OOV rates for newstest$2019$ and the Flores-101 test set are $19$\% and $30$\%, respectively.
It is also plausible to have higher OOV rates for extremely low-resource language pairs with less than $100$K training samples.
Figure~\ref{fig:fig5} indicates the decrease in Romanian-English BLEU score by increasing the rate of OOV words for both word-level and BPE-level models.
Word-level is the model trained with vocabulary set of all actual words without involving any segmentation, while BPE-level models are trained on varying rates of segmentation.

It is worthwhile to mention that in order to ascertain whether additional ``<unk>'' tokens have not the slightest effect on BLEU score, we use the same ``<unk>'' rates and randomly replace them in the training set.
These experiments confirm that the drop in BLEU score is not due to the added ``<unk>'' tokens and it is solely attributable to model failure in translating higher rates of OOV words. 
Based on Figure~\ref{fig:fig5}, the model trained with smaller numbers of BPE merge operations, which splits words into more and shorter segments, is less affected by increasing the OOV rate.
For example, with an OOV rate of $32\%$, comparing with the word-level model, BPE-$10$K line is very close to its baseline without added OOV words.
With a larger number of BPE merge operations, the performance drop increases and gets closer to the performance drop of the word-level model.
Thus, we can conclude that a smaller number of BPE merge operations alleviates the OOV problem. In the next section, we examine this conclusion in more detail.

\begin{figure}[t]
	\centering
	\subfloat[\centering All language pairs]{{\includegraphics[width=0.45\linewidth]{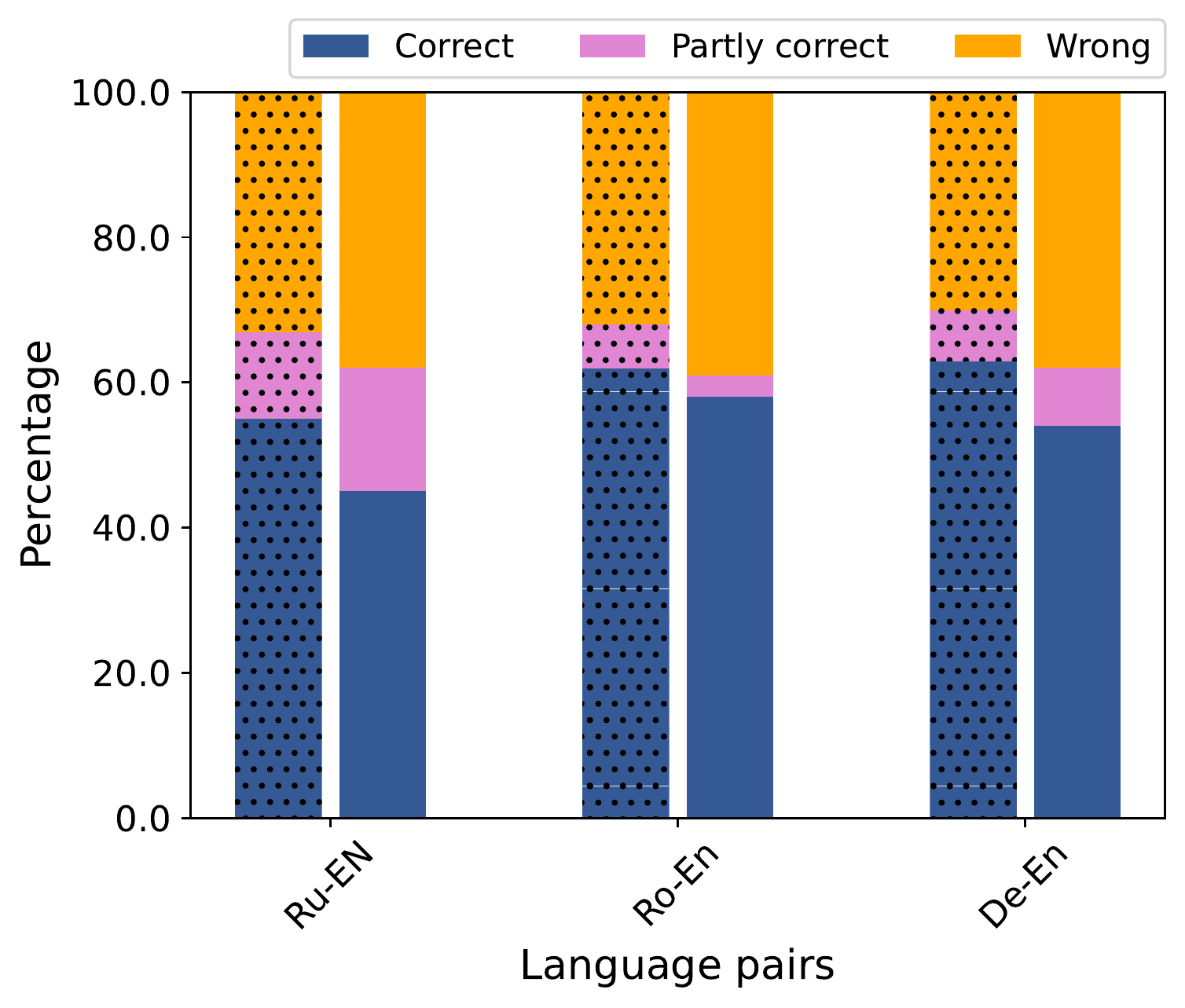} }}%
	\qquad
	\subfloat[\centering Russian-English]{{\includegraphics[width=0.45\linewidth]{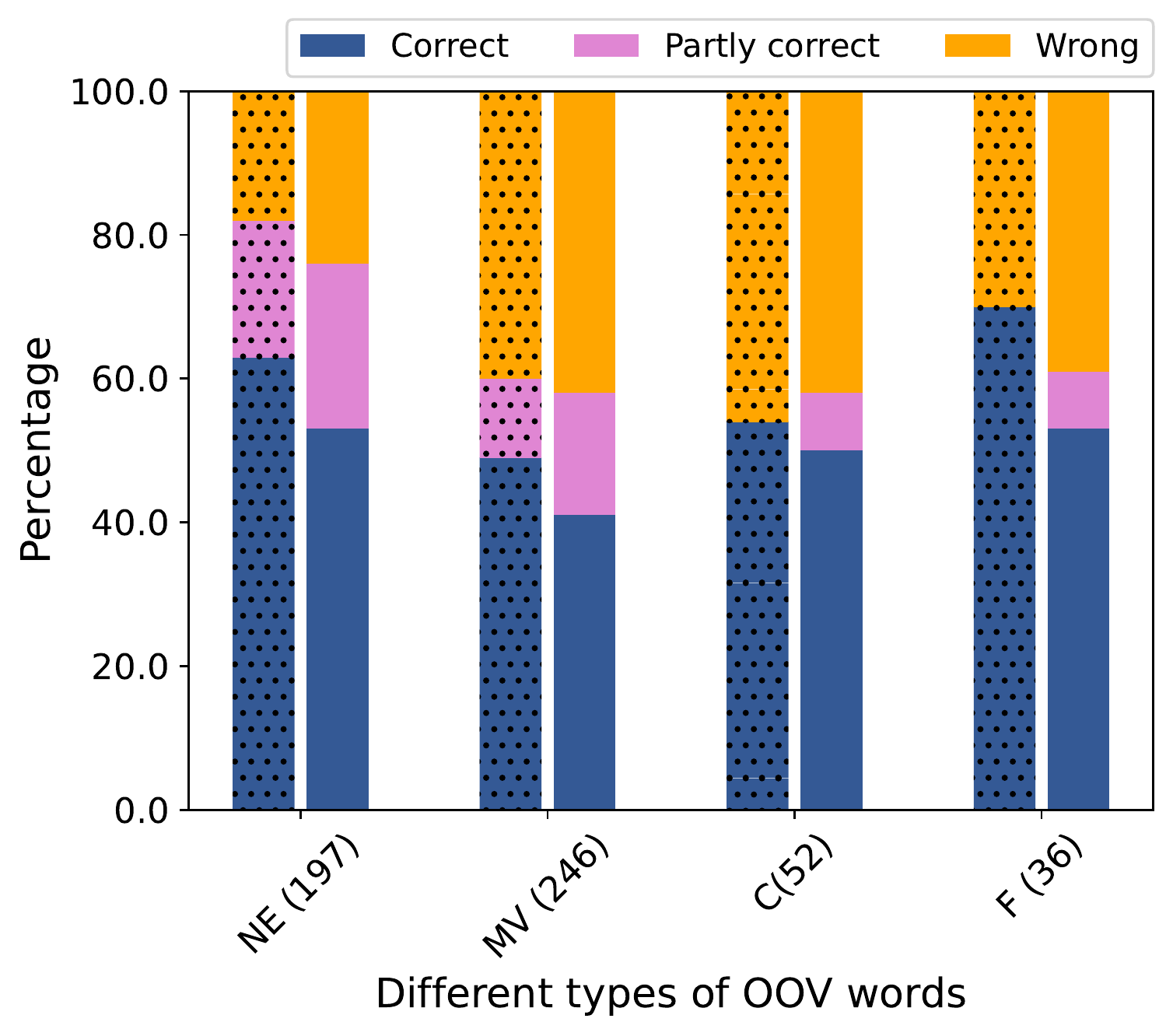} }}%
	\qquad
	\subfloat[\centering Romanian-English]{{\includegraphics[width=0.45\linewidth]{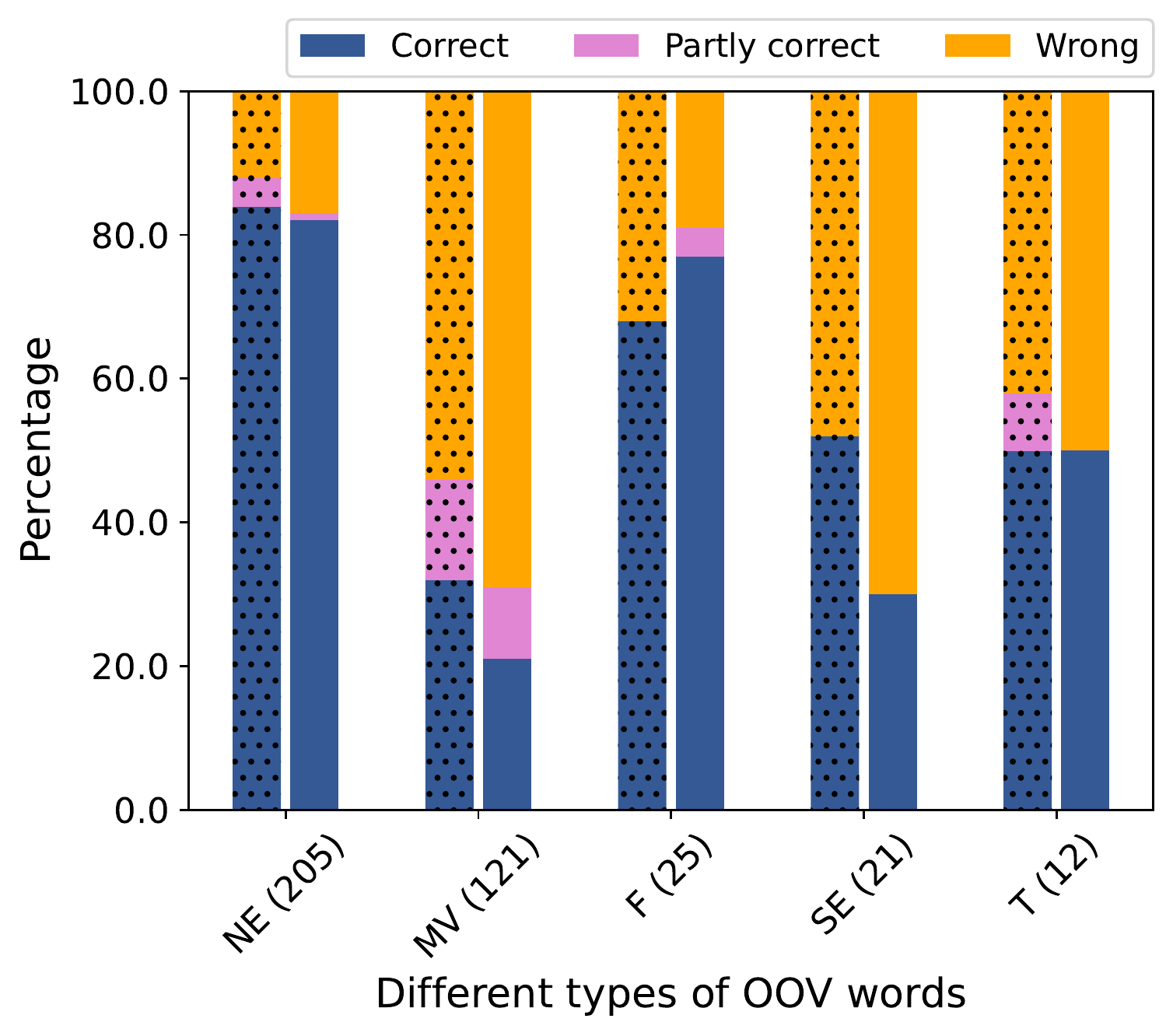} }}%
	\qquad
	\subfloat[\centering German-English]{{\includegraphics[width=0.45\linewidth]{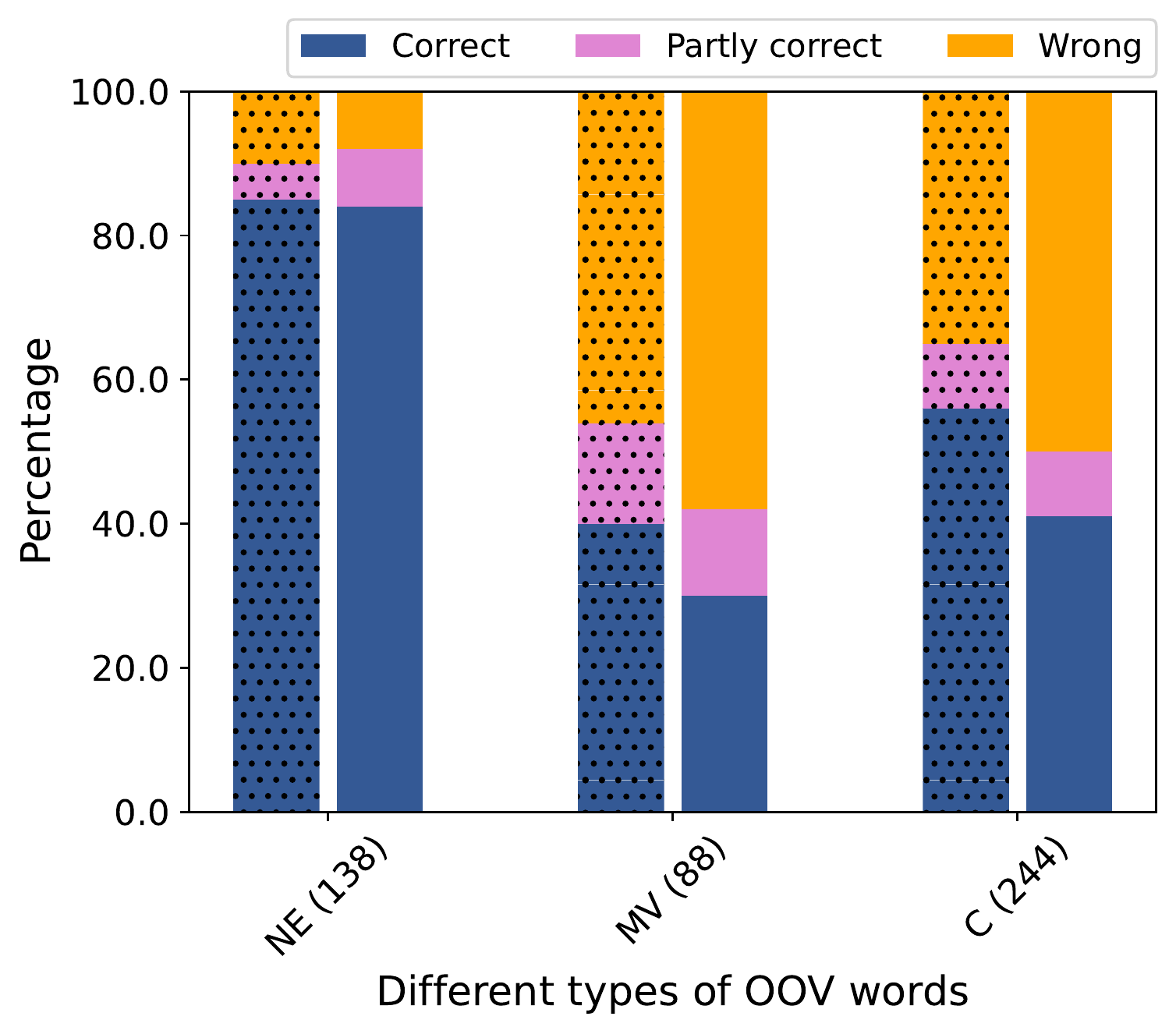} }}%
	\caption{Statistics of translation quality for systems trained with $10$k~(dotted bars indicate BPE-$10$K everywhere) and $37$k BPE merge operations for named-entities~(NE), morphological variants~(MV), compounds~(C), foreign words~(F), spelling errors~(SE), and technical words~(T). Numbers in the parenthesis show the count of OOV type in the corresponding sample. }%
	\label{fig:fig10}%
\end{figure}


\subsection{Translation quality of different OOV types}
\label{sec4.3}
In the previous section, we showed how OOV words still affect the translation quality and using a smaller number of BPE merge operations is presumably more effective to tackle OOV problem.
In this section, we manually analyze the translation quality of OOV words for the systems trained on BPE segmented data with $10$K and $37$K BPE merge operations.
Figure~\ref{fig:fig10}~(a) illustrates the translation quality of OOV words for three language pairs.
Our manual analysis is consistent with Figure~\ref{fig:fig5}, confirming that a smaller number of BPE merge operations is beneficial for translating OOV words.
However,there is an apparent contradiction with Figure~\ref{fig:fig5} showing that a smaller number of BPE merge operations solves the OOV issue of the word-level model, while based on our manual analysis, BPE is only able to translate roughly $60$\% of the OOVs. This contradiction is due to the fact that BLEU tends to neglect local errors~\citep{GuillouHLL18} and the manual assessment is the more precise way to analyze the translation quality of OOV words.

Our preliminary analysis shows that OOV words usually fall into six categories:
named entities~(NE), compounds~(C), morphological variants~(MV), spelling errors~(SE), technical words~(T), and foreign words~(F).  In order to see how well a model trained on BPE segmented data can translate different types of OOV words, we manually label our sample of $400$ OOV words as described in Section~\ref{sec:2.1} for three different language directions. 
For each language pair, we only plot the OOV types with more than $10$ OOVs in the corresponding sample of $400$ OOV types.
As shown in Figure~\ref{fig:fig10}~(b-d), for all language directions, the number of wrong translations is lower for named entities, especially for German-English and Romanian-English presumably due to their high rate of lexical similarity and the same Latin script (except for Romanian declensions).
Morphological variants have the lowest rate of correct translations in all language pairs, which is especially problematic for Russian and Romanian as two morphologically rich languages. Also, for German as a compounding language, only $56$\% of compound OOVs are translated accurately. 
Translation quality of foreign words, spelling errors, and technical words that are very rare compared to the other three OOV types, is moderate to slightly higher for foreign words, as they are mostly English words that are translated to English. In the next sections, we explore some potential reasons for the quality differences of OOV translations. 
\begin{figure}[t]%
	\centering
	\subfloat[\centering Russian-English]{{\includegraphics[width=0.31\linewidth]{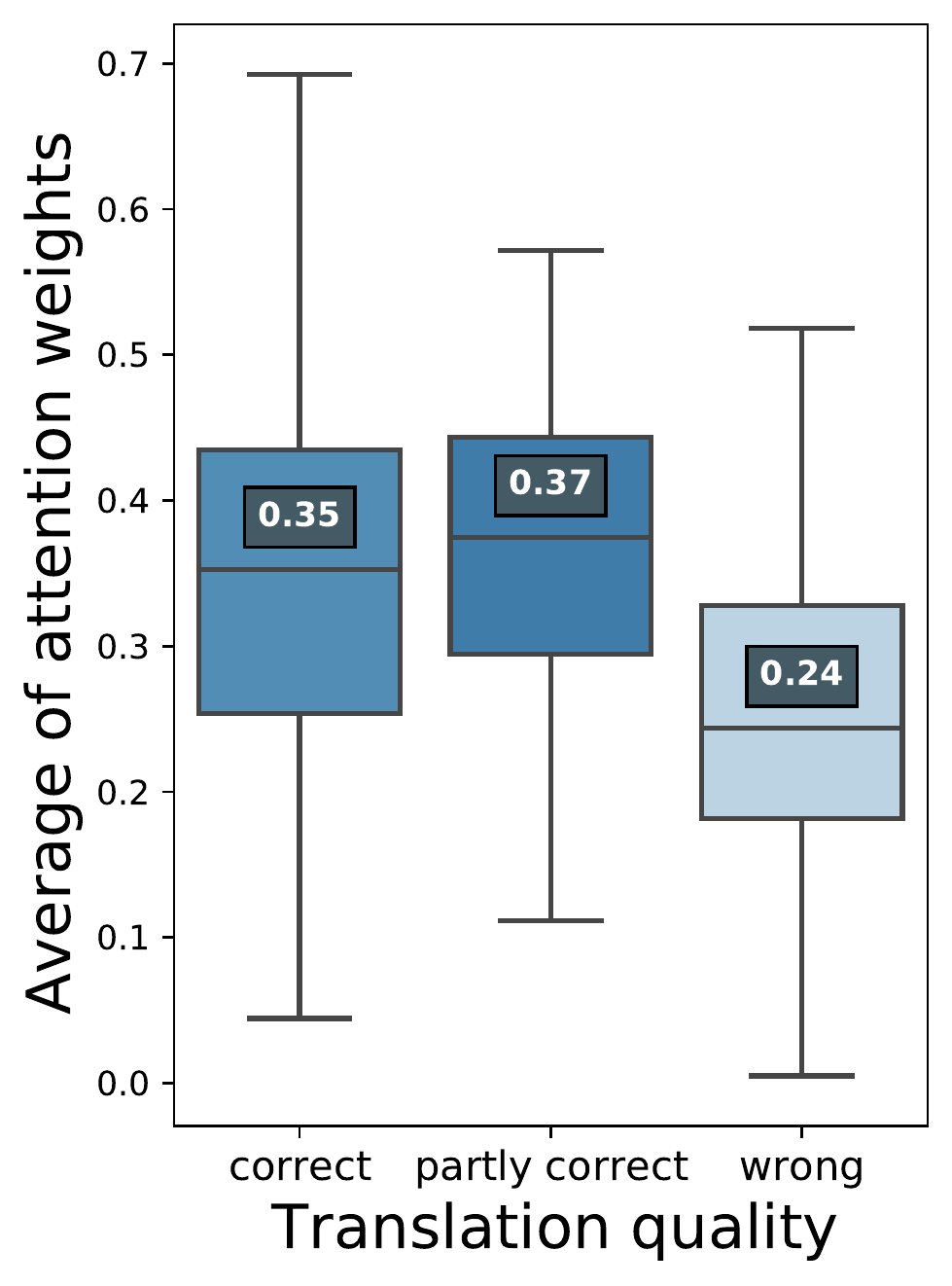} }}%
	\subfloat[\centering Romanian-English]{{\includegraphics[width=0.31\linewidth]{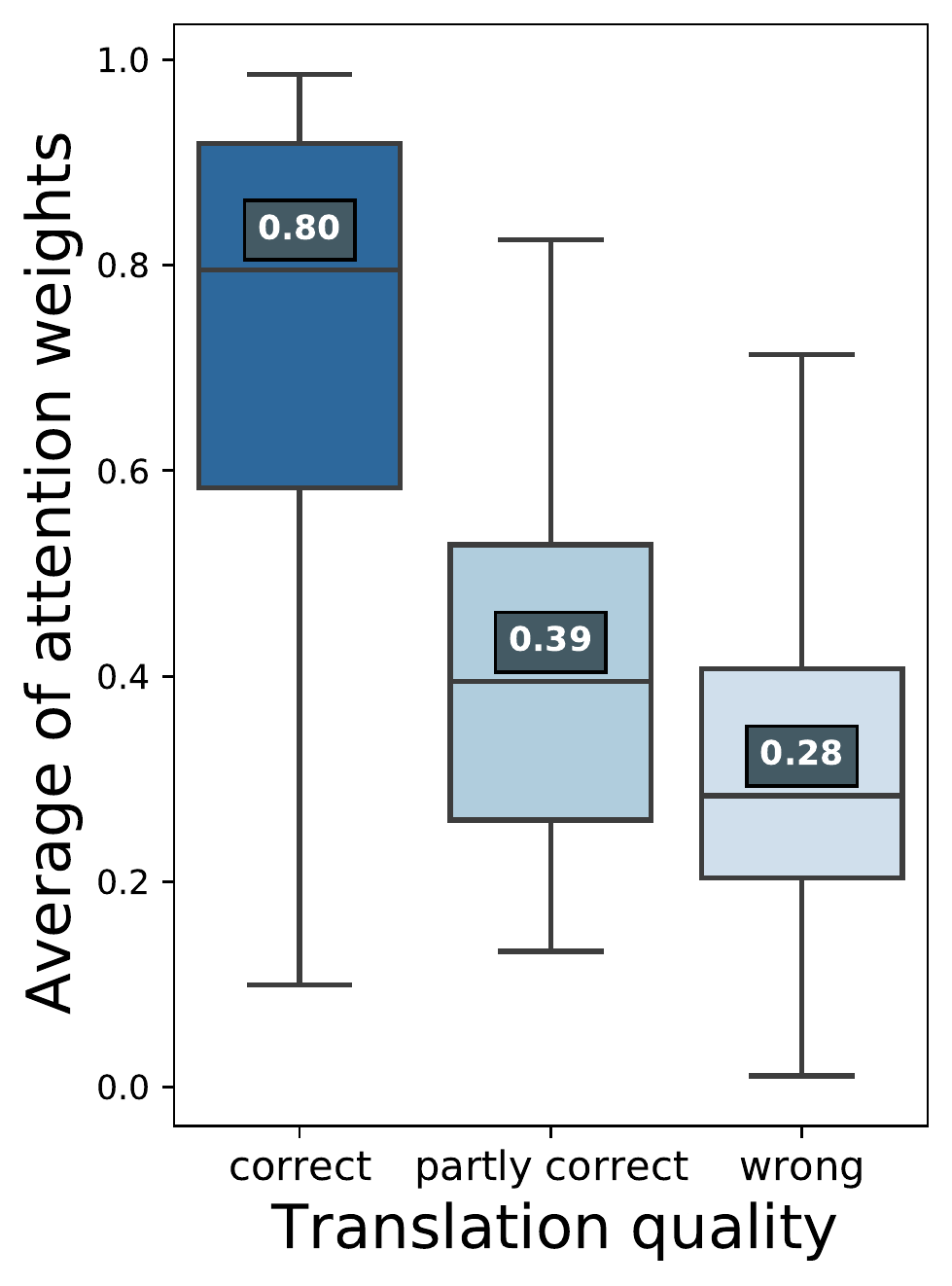} }}%
	\hspace{-.12cm}
	\subfloat[\centering German-English]{{\includegraphics[width=0.31\linewidth]{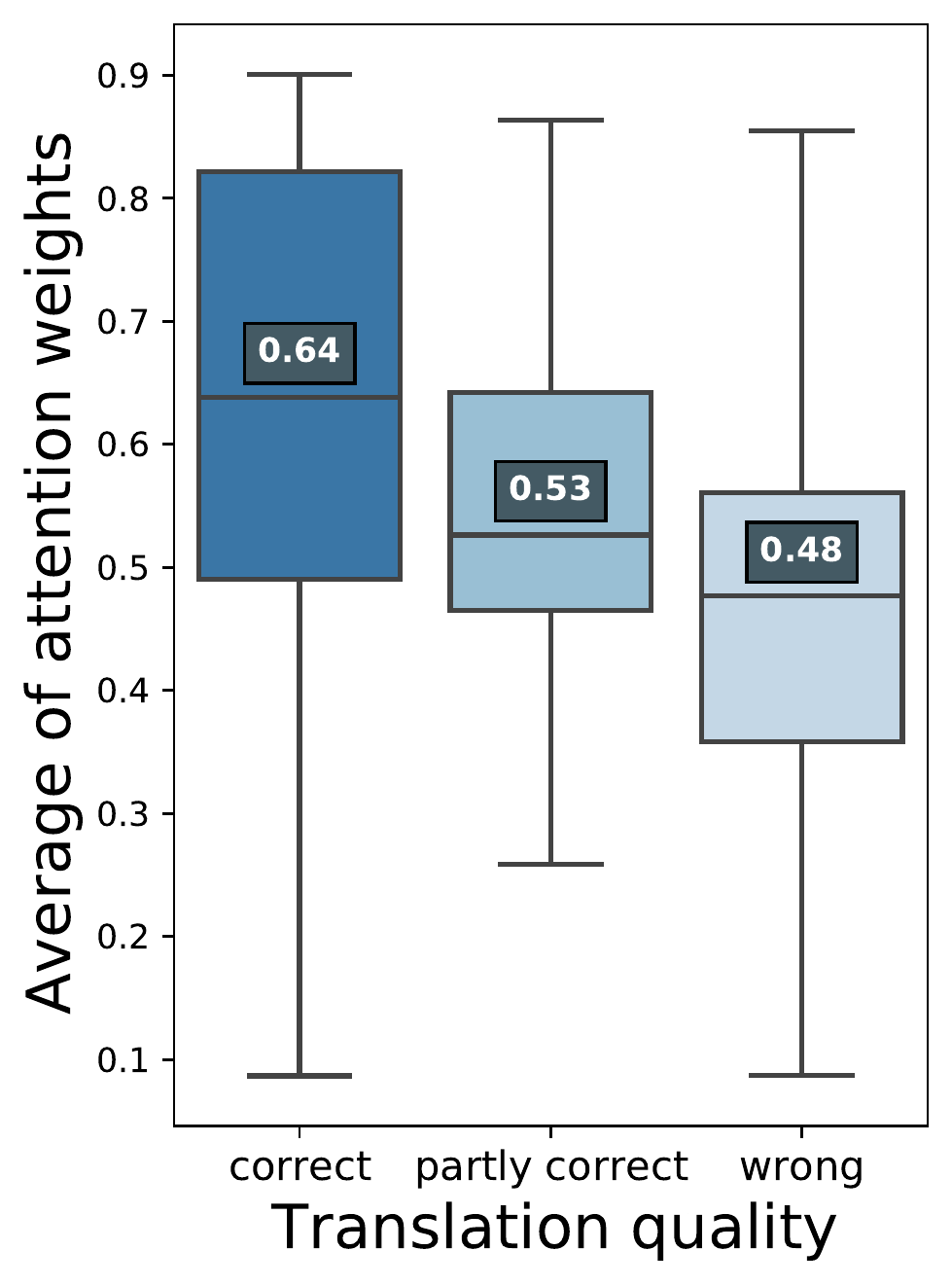} }}%
	\caption{Average of attention weights received by OOV segments from hypothesis token for OOV words with different translation qualities. Each cross-attention weight is computed based on the average weights over heads of the penultimate decoder block. Vertical axis indicates the weight average over the OOV segments. The higher the median, the darker the color.}%
	\label{fig:fig8}%
\end{figure}

\subsection{The amount of attention received by OOV words}
\label{sec4.4}
As mentioned earlier, in order to find the word alignments between source and target sentences,
we use the average over heads of the encoder-decoder attention in the penultimate layer of
Transformer. Specifically,  for each BPE segment in the source language, the target
token with the maximum value of attention weight is identified as the aligned token~\citep{GargPNP19,ChenLCJL20}.
We use this value to explore the amount of attentions received by OOV words.
For this purpose, for each OOV, we take the average over the amounts of attention received by its segments as the amount of attention payed by the corresponding generated segments.
It should be noted that the same also holds for the maximum over segments in addition to average.
Figure~\ref{fig:fig8} indicates that OOV words that are translated accurately have received a significantly higher rate of attention compared to OOV words with wrong translations.
Thus, we hypothesize that the ability of the model to translate segmented OOV words correlates well with the attention received by its constituents.
Also, we observe that correct OOV translations in Romanian-English and German-English receive stronger attention than correct OOV translations in Russian-English.
Furthermore, Figure~\ref{fig:fig10} highlights the limited ability of BPE to facilitate the translation of Russian OOVs into English.
Therefore, we conjecture there is an inverse relationship between the distance of languages involved in the translation and the usefulness of the BPE in translating OOV words.
Another conjecture is that BPE is not a good choice for morphologically-rich languages as Russian. Although, strategies for morphologically driven segmentations fail at consistently improving overall translation quality over BPE~\citep{HuckRF17,abs-1812-08621}, no study is yet to explore the effectiveness of these morphology aware methods with the focus on OOV words.
\begin{figure}[t]%
	\centering
	\subfloat[\centering Russian-English]{{\includegraphics[width=0.31\linewidth]{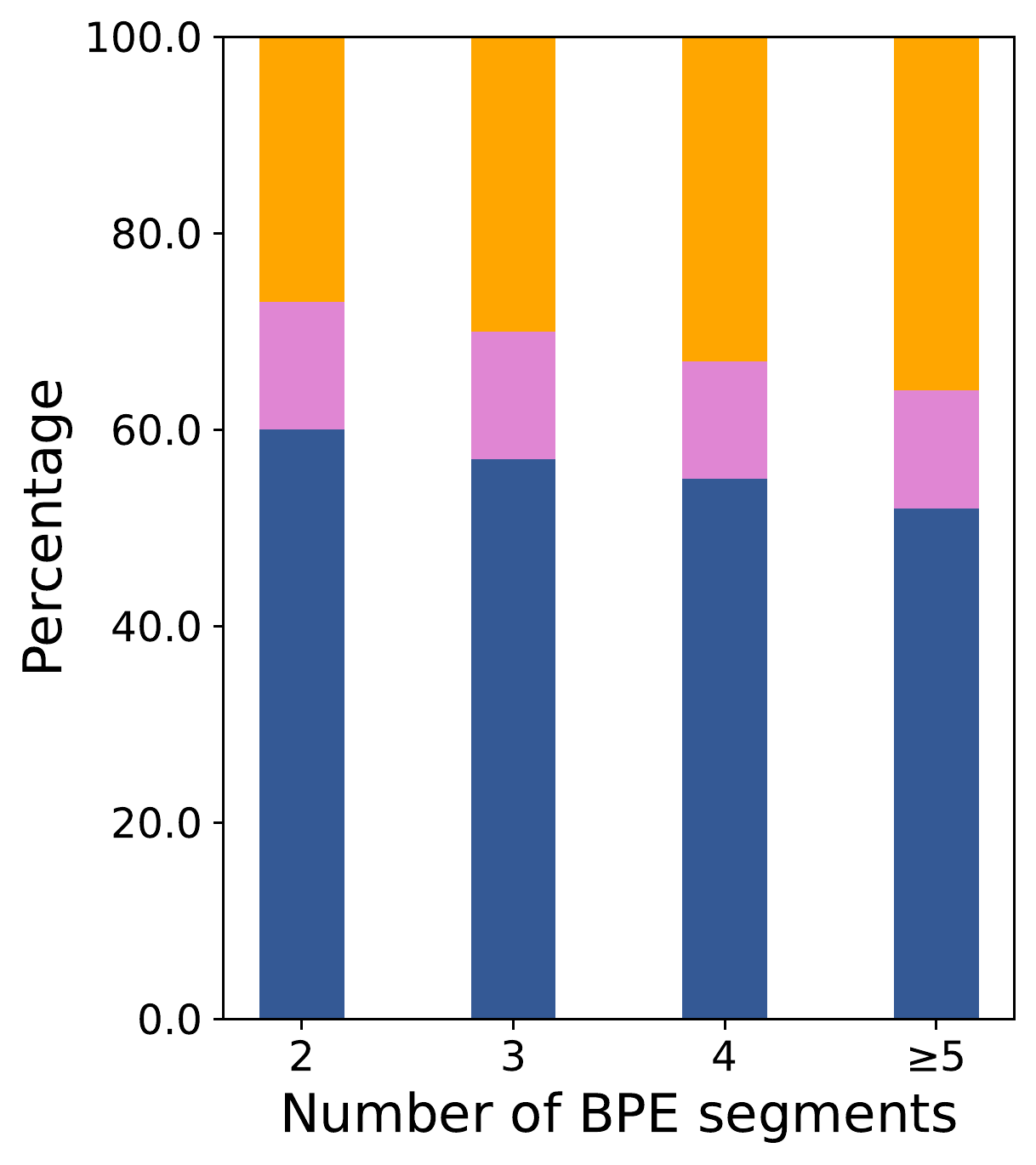} }}%
	\subfloat[\centering Romanian-English]{{\includegraphics[width=0.365\linewidth]{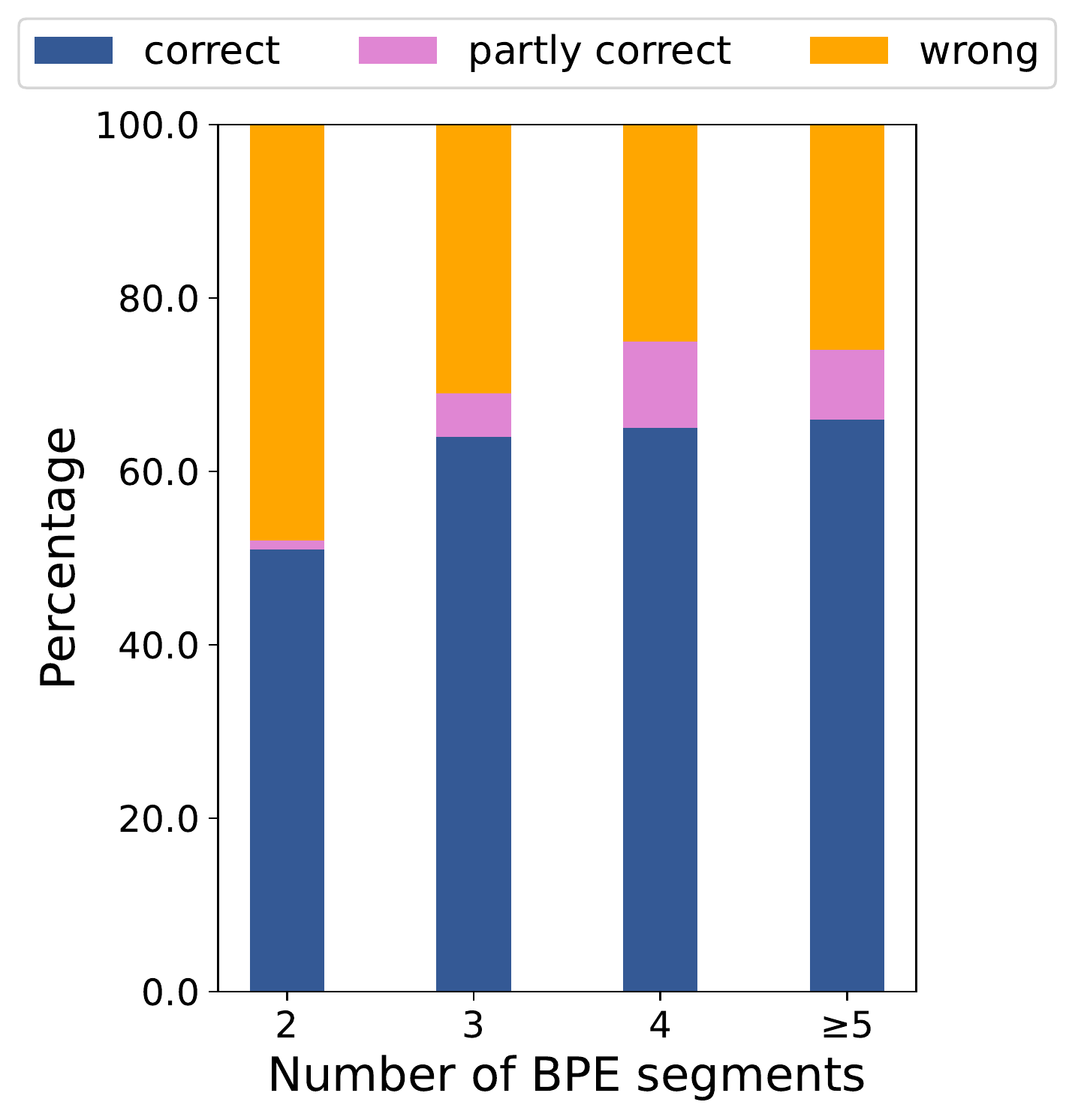} }}%
	\hspace{-.53cm}
	\subfloat[\centering German-English]{{\includegraphics[width=0.31\linewidth]{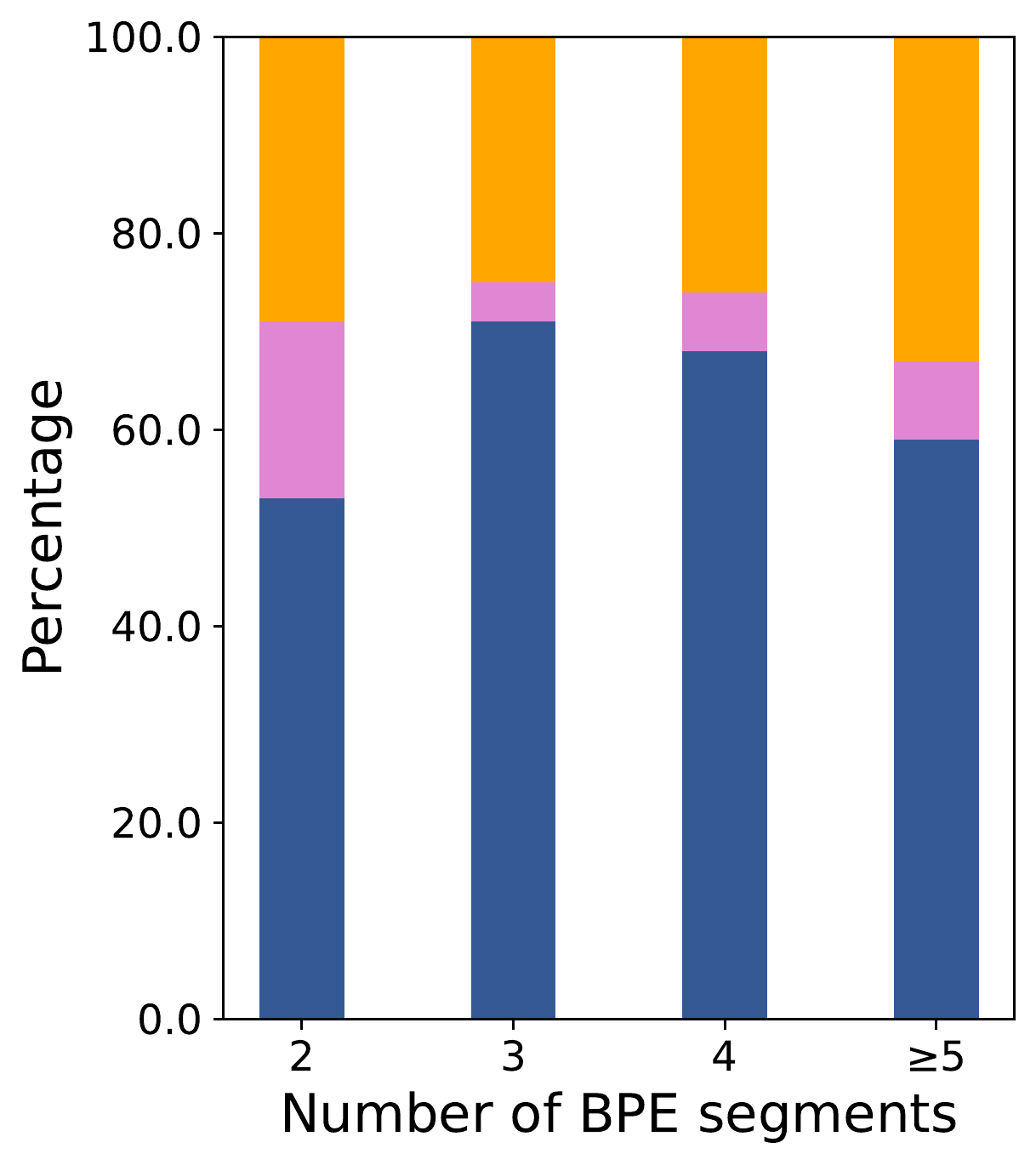} }}%
	\caption{Translation quality for OOV words with different number of BPE segments.}%
	\label{fig:fig3}%
\end{figure}
\begin{figure}[!htb]%
	\centering
	\subfloat[\centering Russian-English]{{\includegraphics[width=0.31\linewidth]{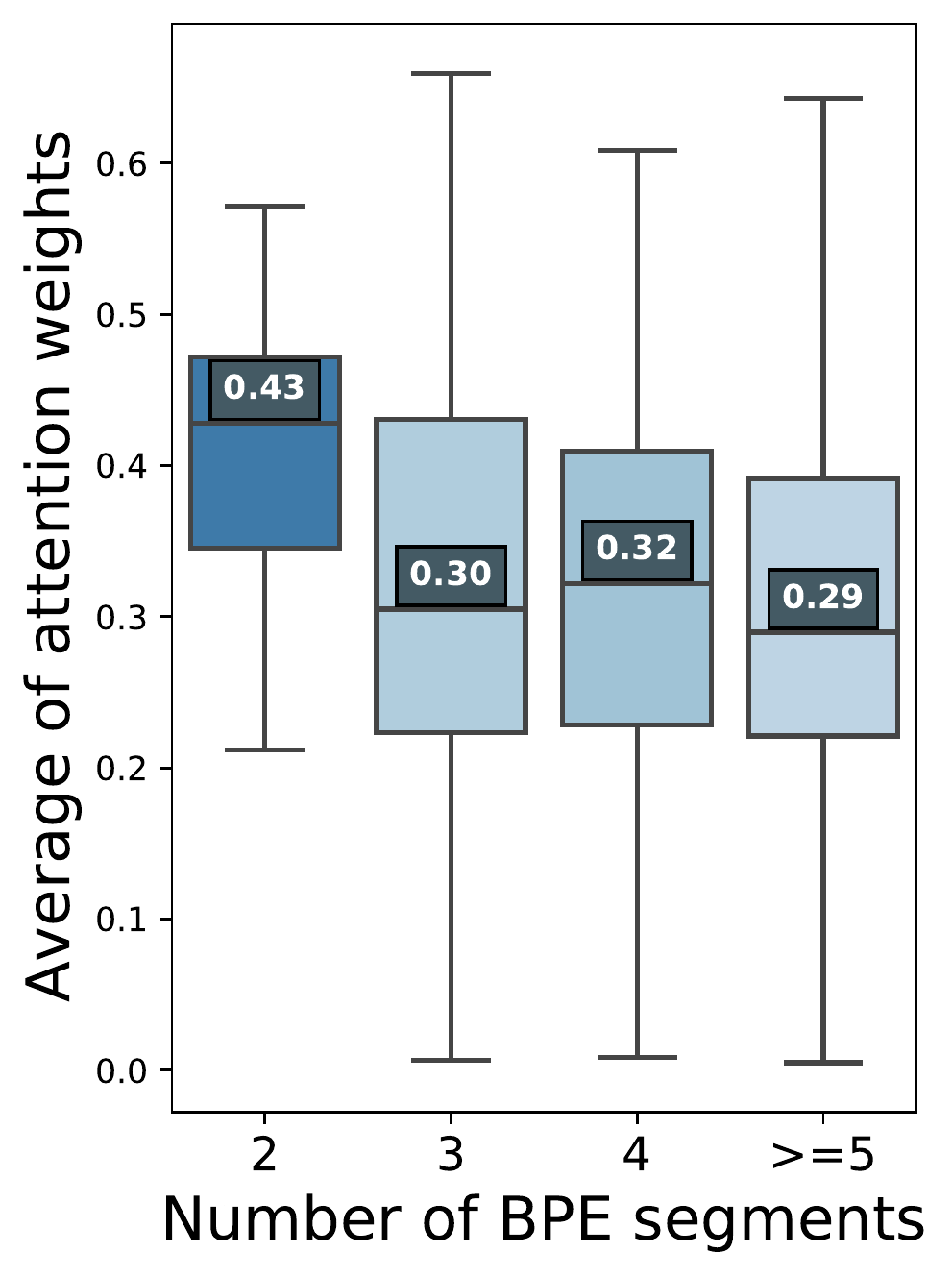} }}%
	\subfloat[\centering Romanian-English]{{\includegraphics[width=0.31\linewidth]{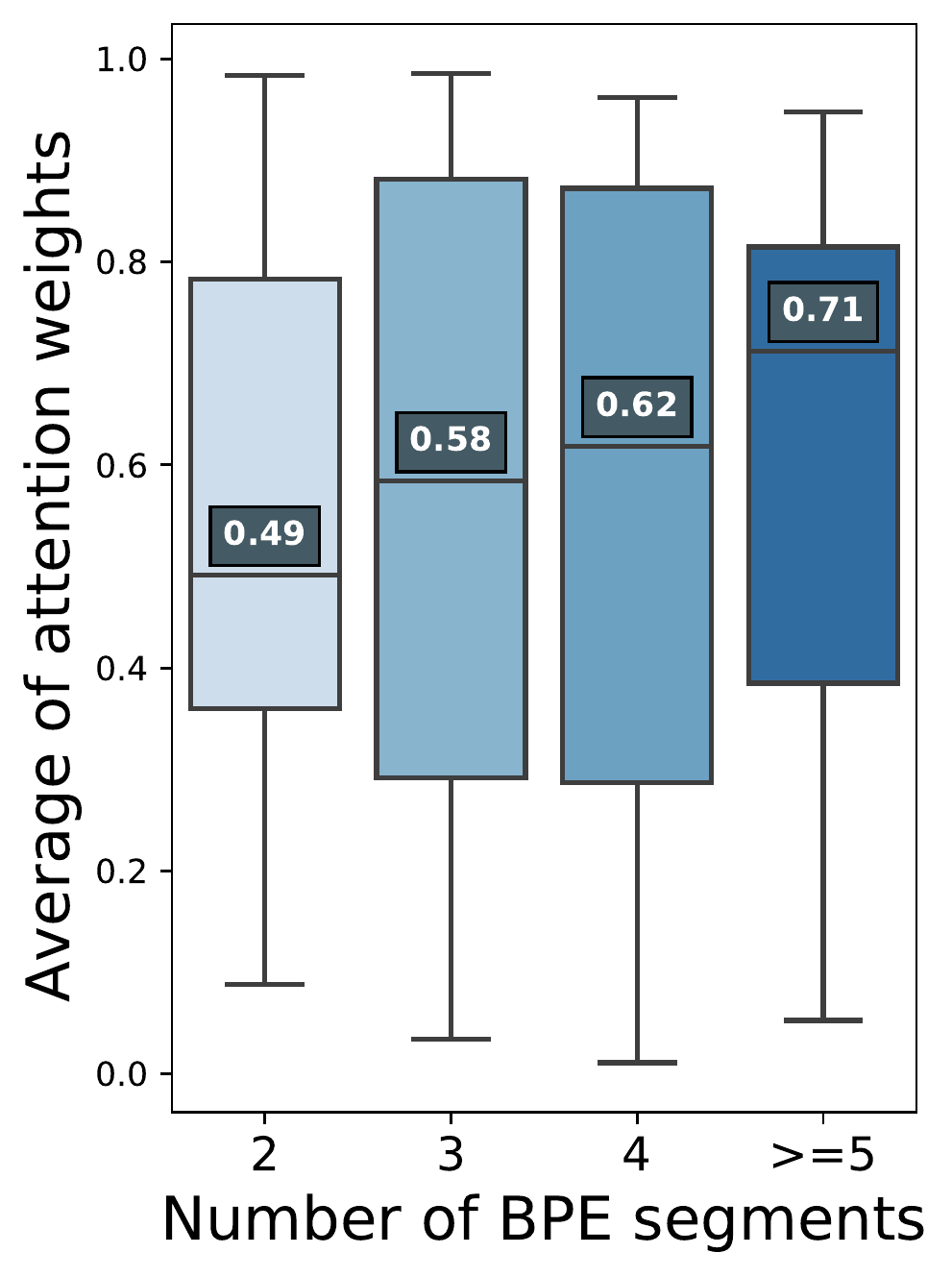} }}%
	\hspace{-.12cm}
	\subfloat[\centering German-English]{{\includegraphics[width=0.31\linewidth]{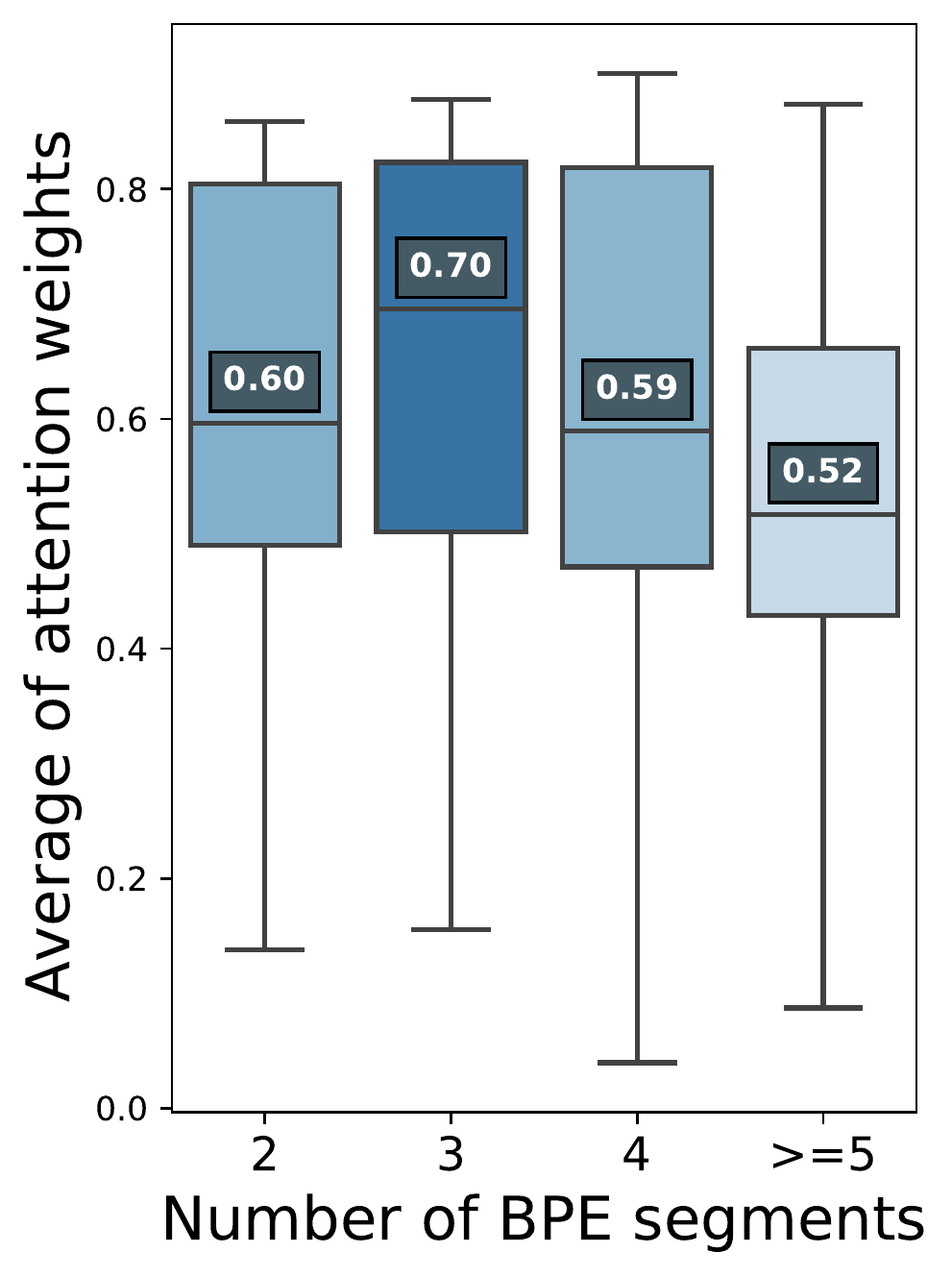} }}%
	\caption{Average of cross-attention weights received by OOV segments from hypothesis token for OOV words with different number of segments. Each cross-attention weight is computed based on the average weights over heads of the penultimate decoder block. Vertical axis indicates the weight average over the OOV segments. The higher the median, the darker the color.}
	\label{fig:fig7}%
\end{figure}

\subsection{Length of BPE segmented OOV word}
BPE keeps the most frequent words intact and splits the rare and unseen words into longer sequences of segments.
In order to scrutinize the relationship between the number of BPE segments for a given OOV word and its translation quality, we use $10$K merge operations, as it is superior in translating BPE segmented OOV words, shown in Section~\ref{sec4.3}.
Figure~\ref{fig:fig3} depicts the quality of OOV translation based on their length. First of all, we find  that  there is no significant correlation between the type of OOV and the length of BPE segmented OOV in each language.
While the shorter lengths seem to have better translations for Russian-English, the opposite is true for Romanian-English. Also, OOV words with a length of $3$ or $4$ segments have a slightly higher rate of correct translations in the case of German-English.
Therefore, we hypothesize splitting OOV words into longer sequences, which is the spirit of BPE, is more effective where there is a higher degree of similarity between language pairs such as Romanian-English and German-English, while having more BPE segments seems to be less effective for Russian-English.
Accordingly we hypothesize more effectiveness of BPE for linguistically similar languages which is consistent with the results of Section~\ref{sec4.4}.
Figure~\ref{fig:fig7} details the attention weights received by OOV words with different lengths which is in complete agreement with Figure~\ref{fig:fig3} showing stronger attention where the length of the OOV words has resulted in higher translation quality.  


\subsection{Effect of frequency of BPE segments in training data}

In this section, we examine if the translation quality is higher where the n-grams of the BPE segments of an OOV word are more frequent in the training data.
We compare the distribution of n-gram frequencies for different quality labels using the one-sided Mann-Whitney U test~\citep{mann1947test}, a non-parametric test to compare the distribution of two groups of data againts each other. Specifically, for unigrams, we compare the frequency distribution of training unigrams that have occurred in correct and partly correct, correct and wrong, and partly correct and wrong translations.
We repeat this for n-grams with $n \in \{1,2,3,4,5\}$. We find that no two distributions are significantly different~$(\alpha=0.1)$.
Thus, there is no evidence in the data to support that the distribution of frequencies of BPE segments for various n-grams are different across different translation qualities.


\subsection{Target-side OOVs}
So far in the paper, the OOV has always referred to the lack of a source-side word in the training vocabulary at inference time. One can also consider the target-side OOV word which is not the purpose of this paper. However, we investigate the relationship between the quality of the translation of a source-side OOV word and the presence of its corresponding reference or correct translation in the vocabulary. In particular, the question is to what extent the translation quality of a given source-side OOV word can be affected when its corresponding correct translation or its corresponding reference in the target side is also an OOV? In our exploration, we observe that for a significant number of correct translations, the reference or the correct output of the model is not present in the training set, which highlights the model ability to generate target-side OOV words. For German-English and Russian-English the number of correct source-side OOV translations with the target-side OOV words is higher than the correct translations for the target-side non-OOV words. However, Romanian-English is vice-versa. Therefore, there is not a consistent behaviour in all language pairs to support that the target-side OOV has a negative effect on the translation of the source-side OOV.

\section{Conclusion}
In this paper, we analyze the translation quality of OOV words in BPE segmented datasets.
Our analysis shows that while BPE has brought significant improvements to NMT in terms of automatic evaluation metrics, the translation quality still suffers from OOV words.
Our experiments show that splitting OOV words into subwords is more effective where there is higher degree of language similarity.
Also, there is a strong correlation between the translation quality and the amount of attention received by OOV words.
On the other hand, there is no evidence to support that the translation quality is dependent on the frequency of BPE segment n-grams in the training data.
Moreover, we find that the translation quality is better for named entity OOV words compared to other word types, especially for language pairs with more lexical similarity.
Furthermore, we showed that automatic evaluation metrics such as BLEU are not able to capture the effectiveness of a word segmentation method for translations of OOVs. Therefore, manual analysis on the translation quality of OOV words is essential to compare different approaches, although it needs annotators in each language and it is very laborious.
In future work, we compare suggested approaches for morphologically-rich languages at the word level.

\section{Acknowledgements}

This research was funded in part by the Netherlands Organization for Scientific Research (NWO) under project number VI.C.192.080. We thank all the members of the UvA Language Technology Lab for their constant feedback on our work. Special thanks to the following people for their support, without whose help and annotation this work would never have been possible: Evgeniia Tokarchuk, Kata Naszadi, Maria Heuss, Mozhdeh Ariannezhad, and Vera Provatorova. Finally, we want to
thank the anonymous reviewers for their valuable
input and suggestions.

\bibliographystyle{apalike}
\bibliography{main}

\end{document}